\documentclass{article}

\usepackage{arxiv}

\usepackage[utf8]{inputenc} % allow utf-8 input
\usepackage[T1]{fontenc}    % use 8-bit T1 fonts
\usepackage{hyperref}       % hyperlinks
\usepackage{url}            % simple URL typesetting
\usepackage{booktabs}       % professional-quality tables
\usepackage{amsfonts}       % blackboard math symbols
\usepackage{nicefrac}       % compact symbols for 1/2, etc.
\usepackage{microtype}      % microtypography
\usepackage{lipsum}		    % Can be removed after putting your text content
\usepackage{graphicx}
\usepackage{natbib}
\usepackage{doi}

\setcitestyle{numbers,square} 

\title{Robust Multi-Disease Retinal Classification via Xception-Based Transfer Learning and W-Net Vessel Segmentation}

\date{September 9, 2020}	% Here you can change the date presented in the paper title
\author{ 
    Mohammad Sadegh Gholizadeh \\
	Computer Engineering Department\\
	Shahid Rajaee University\\
	\texttt{gholizadeh@sru.ac.ir} \\
	\And
	Amir Arsalan Rezapour \\
	Computer Engineering Department\\
	Shahid Rajaee University\\
	\texttt{arsalanrzp@sru.ac.ir} \\
}

\renewcommand{\shorttitle}{Robust Multi-Disease Retinal Classification}

\hypersetup{
pdftitle={Robust Multi-Disease Retinal Classification via Xception-Based Transfer Learning and W-Net Vessel Segmentation},
pdfsubject={cs.CV, cs.AI},
pdfauthor={Mohammad Sadegh Gholizadeh, Amir Arsalan Rezapour},
pdfkeywords={Ocular disease classification, semantic segmentation, deep learning, human-in-the-loop AI},
}

\begin{document}
\maketitle

% --- ABSTRACT ---
\begin{abstract}
In recent years, the incidence of vision-threatening eye diseases has risen dramatically, necessitating scalable and accurate screening solutions. This paper presents a comprehensive study on deep learning architectures for the automated diagnosis of ocular conditions. To mitigate the "black-box" limitations of standard convolutional neural networks (CNNs), we implement a pipeline that combines deep feature extraction with interpretable image processing modules. Specifically, we focus on high-fidelity retinal vessel segmentation as an auxiliary task to guide the classification process. By grounding the model's predictions in clinically relevant morphological features, we aim to bridge the gap between algorithmic output and expert medical validation, thereby reducing false positives and improving deployment viability in clinical settings.
\end{abstract}

% --- KEYWORDS ---
% Note: The template uses \and to separate keywords

% ... The rest of the paper will go here ...

\section{Introduction}
The escalating prevalence of ocular pathologies is driven largely by demographic aging and the prolonged exposure to digital screens characteristic of modern lifestyles. Timely diagnosis and appropriate therapeutic intervention are critical for preventing irreversible vision loss and preserving patient quality of life. However, conventional diagnostic workflows rely heavily on the subjective expertise of clinicians. This dependence introduces significant inter-observer variability and increases the risk of misdiagnosis, potentially leading to adverse patient outcomes.

The intersection of ophthalmology and Artificial Intelligence (AI) promises a paradigm shift in diagnostic protocols. Deep learning methodologies have already demonstrated remarkable success in medical imaging tasks, ranging from oncology and cardiology to neurodegenerative disorders like Alzheimer's disease. In the context of ocular health, we categorize the computational landscape into three primary objectives: (1) automated disease detection, (2) severity grading, and (3) semantic segmentation of clinically relevant biomarkers (e.g., retinal vessels) to support physician interpretation.

This paper addresses these challenges by developing a robust deep learning framework for retinal analysis. The remainder of this paper is organized as follows: Section II provides an overview of common ocular pathologies and the datasets utilized. Section III reviews prior art and state-of-the-art methodologies. Finally, we detail our proposed architecture and segmentation pipeline, followed by a comprehensive analysis of the empirical results.

\section{Clinical Background: Ocular Pathologies}

Automated diagnostic systems must distinguish between subtle morphological biomarkers characteristic of specific ocular conditions. Here, we briefly characterize the six pathologies targeted in this study, emphasizing the visual features relevant to classification.

\begin{figure}[t]
	\centering
	\includegraphics[width=0.9\linewidth]{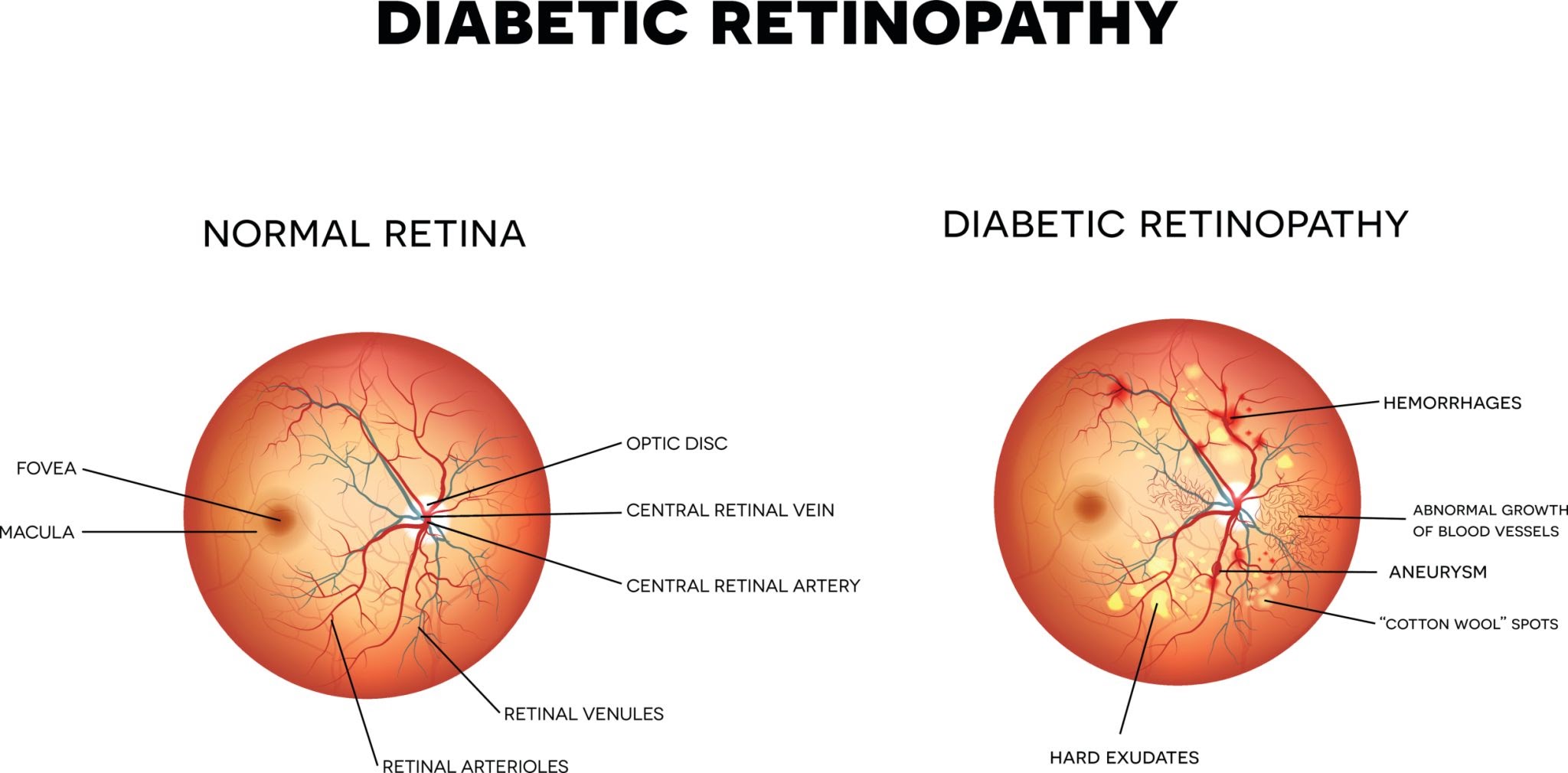}
	\caption{Comparison of a normal retina versus one affected by Diabetic Retinopathy, highlighting key biomarkers such as hemorrhages, aneurysms, and hard exudates.}
	\label{fig:dr_diagram}
\end{figure}

\textbf{Retinal and Vascular Pathologies.} 
\textbf{Diabetic Retinopathy (DR)} is a microvascular complication of chronic hyperglycemia, triggering neovascularization in response to ischemia. These fragile new vessels frequently rupture, causing hemorrhages and fluid leakage that obscure retinal features (see Figure \ref{fig:dr_diagram}). Sharing a similar vascular etiology, \textbf{Hypertensive Retinopathy} results from systemic hypertension; it induces arteriosclerotic changes, vascular leakage, and focal ischemia (manifesting as cotton-wool spots). Distinct from vascular damage, \textbf{Age-Related Macular Degeneration (AMD)} involves the deterioration of the macula, characterized by the accumulation of extracellular drusen beneath the retinal pigment epithelium and central visual field loss.

\begin{figure}[htbp]
    \centering
    \begin{minipage}{0.48\textwidth}
        \centering
        \includegraphics[width=0.9\linewidth]{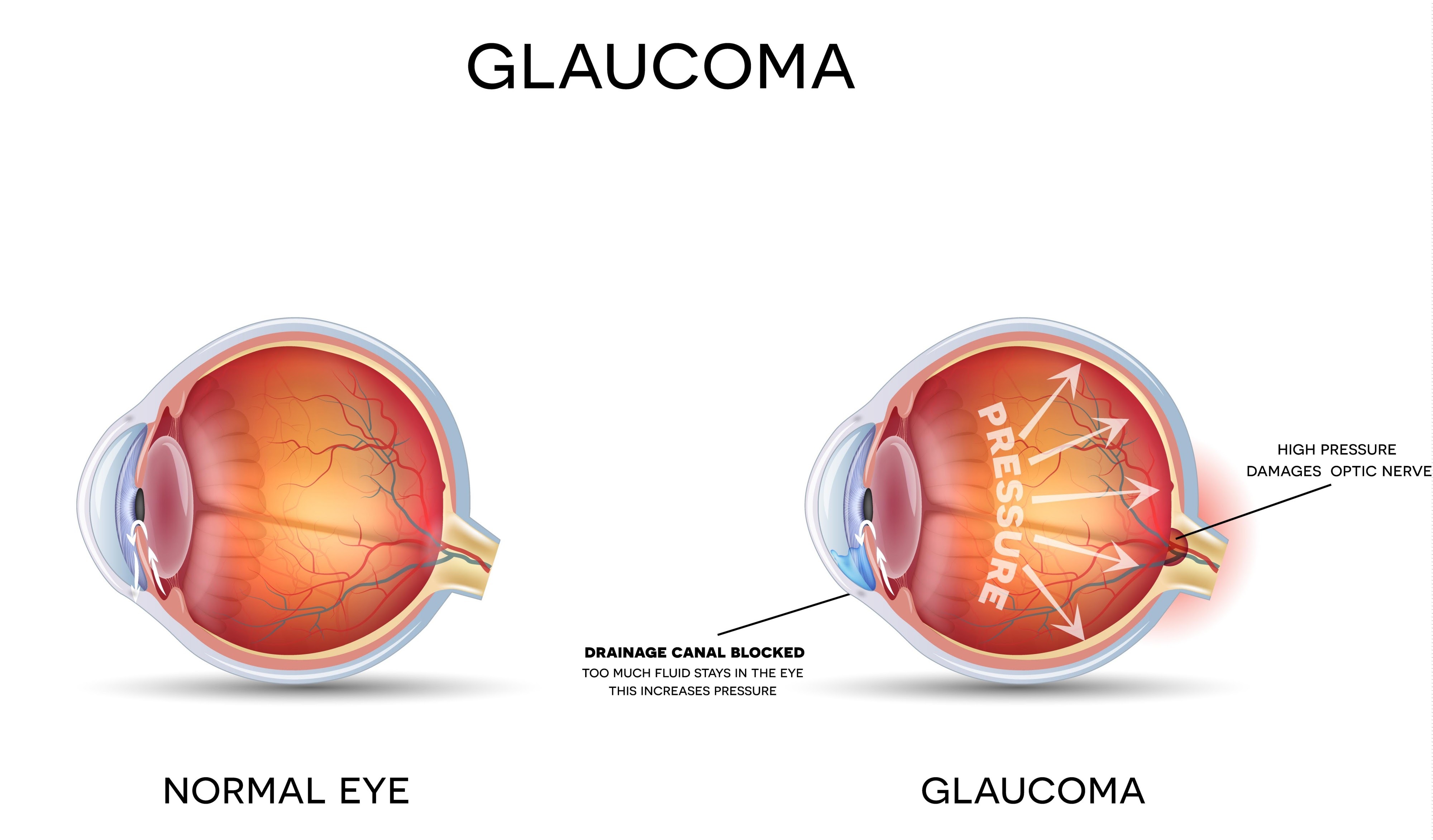}
        \caption{Cross-section view illustrating the lens opacification characteristic of Cataracts.}
        \label{fig:cataract_diagram}
    \end{minipage}\hfill
    \begin{minipage}{0.48\textwidth}
        \centering
        \includegraphics[width=0.9\linewidth]{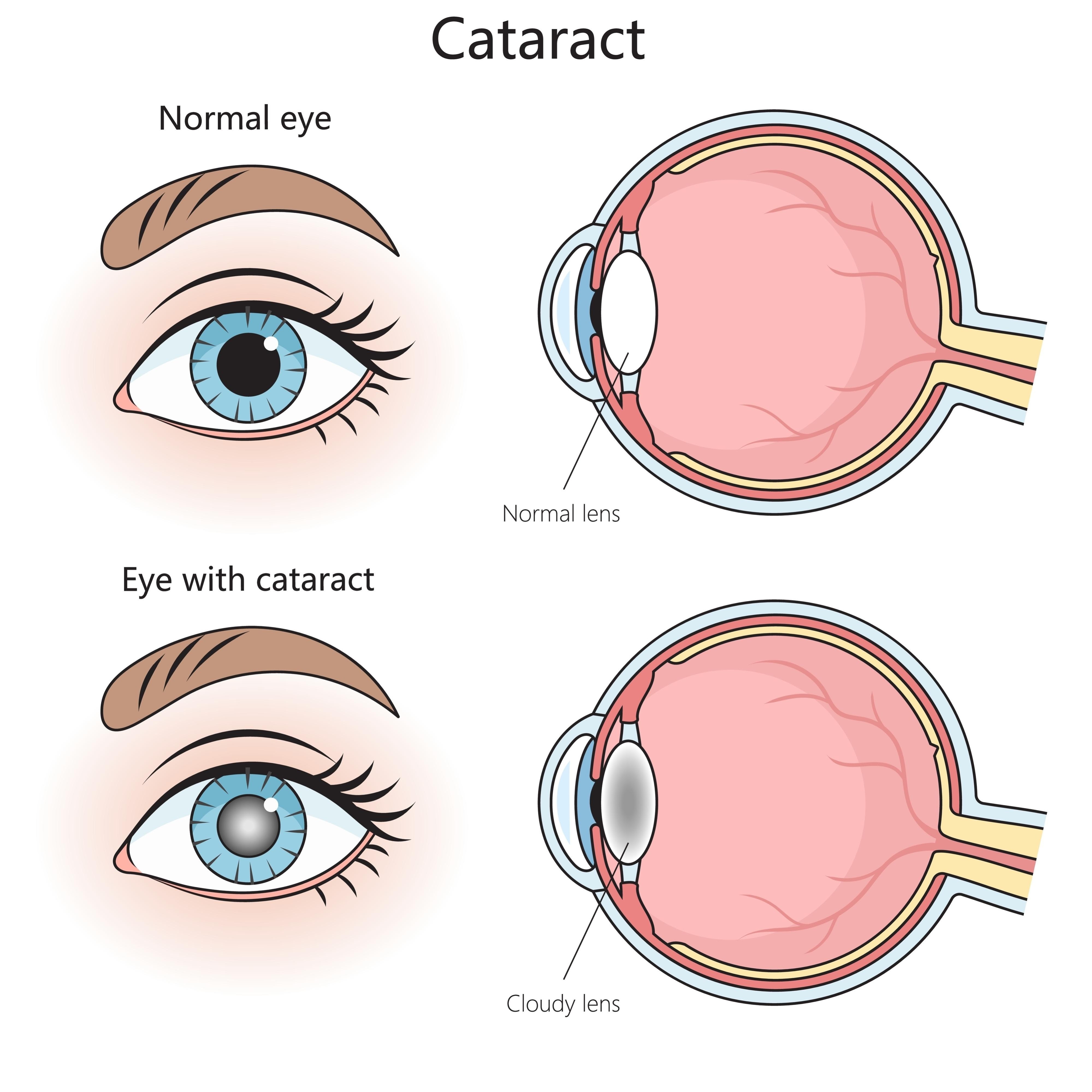}
        \caption{Glaucoma mechanism showing IOP elevation and optic nerve damage.}
        \label{fig:glaucoma_diagram}
    \end{minipage}
\end{figure}

\textbf{Structural and Optic Neuropathies.} 
\textbf{Glaucoma} comprises a group of neuropathies driven by elevated intraocular pressure (IOP), which induces progressive degeneration of the optic nerve head (ONH) and irreversible peripheral vision loss (Figure \ref{fig:glaucoma_diagram}). In contrast, \textbf{Cataracts} involve the opacification of the crystalline lens, obstructing light transmission and blurring the visual input globally (Figure \ref{fig:cataract_diagram}). Finally, \textbf{Pathologic Myopia} is defined by excessive axial elongation of the eyeball, leading to refractive errors and structural deformations that risk chorioretinal atrophy.

\textbf{Epidemiological Significance.} 
These conditions represent leading causes of preventable blindness. Globally, approximately 400 million individuals suffer from diabetes, with one-third developing DR. In Iran specifically, a 10\% diabetes prevalence combined with a rapidly aging demographic projects a significant rise in the incidence of these disorders \citep{ref1}.

% [Insert Figures Here - Grouping them is often preferred in 2-column formats]

\section{Datasets and Modalities}
The availability of large-scale, open-access ophthalmic repositories is a cornerstone of modern medical computer vision. In this study, we leverage datasets spanning multiple imaging modalities to ensure robust model generalization. A comprehensive breakdown of these sources is provided in the supplementary material; here, we detail the primary modalities and the specific subsets utilized.

\subsection{Imaging Modalities}

\textbf{Color Fundus Photography (CFP):} CFP is the most prevalent modality for screening ocular pathology. It involves capturing a 2D projection of the 3D retinal tissue using reflected light. To ensure visualization of the peripheral retina, the procedure typically requires pharmacological mydriasis (pupil dilation). CFP is the standard for diagnosing diabetic retinopathy, glaucoma, and cataracts.

\textbf{Optical Coherence Tomography (OCT):} OCT is a non-invasive imaging technique that utilizes low-coherence interferometry to capture high-resolution cross-sections of the retina. Unlike fundus photography, OCT penetrates retinal layers, allowing for the precise measurement of retinal thickness and the identification of intra-retinal pathologies such as drusen or edema.

\subsection{Data Sources}
We utilized the following benchmark datasets for training and evaluation:

\begin{itemize}
    \item \textbf{Multi-Disease Fundus Dataset (ODIR-5K):} A structured database comprising color fundus images from 3,500 patients. It includes annotations for both left and right eyes, covering conditions such as cataracts, diabetes, and glaucoma. Metadata is limited to disease labels, ensuring models rely solely on visual features rather than demographic biases.
    
    \item \textbf{Retinal OCT Dataset (Kermany et al.):} A large-scale corpus containing 84,450 spectral-domain OCT images. The dataset is stratified into training, validation, and test sets and labeled for four classes: Choroidal Neovascularization (CNV), Drusen, Diabetic Macular Edema (DME), and Normal.
    
    \item \textbf{Glaucoma Segmentation Dataset:} A dedicated subset focused on optic disc assessment, consisting of 520 training images and 130 validation images. This dataset is restricted to binary glaucoma labels, facilitating targeted optic nerve head analysis.
    
    \item \textbf{Multimodal Paired Dataset:} A collection containing paired OCT and Fundus images, essential for tasks involving cross-modality synthesis or domain adaptation.
\end{itemize}

\section{Related Work}
Current methodologies in ophthalmic image analysis can be broadly categorized into traditional machine learning approaches and Deep Learning (DL) frameworks. While classical methods such as Support Vector Machines (SVM) and Random Forests have historically served as baselines, the availability of large-scale datasets has shifted the state-of-the-art toward Convolutional Neural Networks (CNNs).

\subsection{Diabetic Retinopathy (DR) Diagnosis}
Recent work has demonstrated the efficacy of hybrid architectures for DR detection. For instance, \citet{ref2} proposed a pipeline that combines deep feature extraction with a decision tree classifier. Their approach emphasizes the importance of preprocessing—specifically pixel normalization and Region of Interest (ROI) cropping—and robust data augmentation to mitigate class imbalance. Moving beyond global classification, \citet{ref3} focused on the detection of specific biomarkers, utilizing a deep network to identify pathological features initially labeled by experts. However, semi-automated systems often face scalability challenges. The Deep Fusion Classification Network proposed in \citet{ref4} achieves high accuracy by analyzing twelve distinct retinal layers extracted from OCT imagery. Despite its performance, this method relies on manual segmentation of layers, limiting its utility in fully automated clinical workflows.

\subsection{Glaucoma Detection \& ROI Analysis}
In glaucoma screening, the optic disc is a critical morphological feature. Research by \citet{ref5} and \citet{ref6} indicates that networks focusing specifically on the optic disc ROI yield superior performance compared to those analyzing the entire global fundus image. To ensure generalization across diverse acquisition devices, \citet{ref7} introduced a cross-dataset validation strategy, training on four disparate datasets while testing on a fifth, thereby benchmarking model robustness against domain shifts.

\subsection{Semantic Segmentation}
Accurate segmentation of retinal vasculature is essential for calculating clinical indices such as the artery-to-vein ratio. While the U-Net architecture remains the standard for medical segmentation, \citet{ref8} introduced the "W-Net," a cascade of two U-Nets connected in series. This modification enhances the segmentation of fine vessel structures, critical for blood flow analysis.

\section{Proposed Methodology}
Based on the comparative analysis of related work, we adopt a Deep Learning (DL) approach for the automated classification of ocular pathologies, leveraging its superior feature representation capabilities over traditional machine learning. Our proposed framework is bifurcated into two distinct phases designed to optimize both diagnostic accuracy and clinical utility:

\begin{description}
    \item[High-Precision Classification:] The primary module focuses on minimizing diagnostic error across multi-label fundus imagery using advanced Convolutional Neural Networks (CNNs).
    
    \item[Clinical Decision Support \& Interpretability:] To bridge the gap between "black-box" predictions and medical experts, the second phase provides contextual evidence. This includes Content-Based Image Retrieval (CBIR) to present clinicians with historically similar cases, and Visual Attribution (Saliency Mapping/Segmentation) to highlight pathological regions of interest, thereby facilitating explainable diagnosis.
\end{description}

\section{Methodology: Automated Disease Detection}
This section details the proposed framework for multi-label ocular disease classification. Our pipeline consists of three stages: data curation and balancing, robust image preprocessing, and a hybrid classification architecture utilizing transfer learning.

\subsection{Dataset Curation and Distribution}
We utilized Color Fundus Photography (CFP) as the primary modality due to its clinical ubiquity and the availability of large-scale public benchmarks. To construct a balanced training corpus, we aggregated data from multiple sources, as summarized in Table \ref{tab:datasets}.

\paragraph{1. Primary Source (ODIR-5K)}
The Ocular Disease Intelligent Recognition (ODIR) dataset serves as the backbone of our study. It contains 7,000 fundus images (left and right eyes) annotated with eight labels: Normal, Diabetes, Glaucoma, Cataract, AMD, Hypertension, Myopia, and "Other."

\paragraph{2. Class Balancing (Cataract Augmentation)}
An initial analysis of ODIR revealed significant class imbalance, particularly for the 'Cataract' class. To mitigate bias and improve minority class recall, we augmented our training set with the Cataract Dataset (Syed Ali et al.), which provided an additional 600 samples across four categories.

\begin{table}[htbp]
\caption{Summary of Utilized Datasets and Modalities}
\label{tab:datasets}
\centering
\resizebox{\textwidth}{!}{% Resize table to fit text width
\begin{tabular}{llccc}
\toprule
\textbf{Dataset Name} & \textbf{Pathology Labels} & \textbf{Modality} & \textbf{Samples} & \textbf{Patients} \\
\midrule
ODIR-5K & Multi-label (8 classes) & Fundus & 6,392 & 3,500 \\
Retinal OCT (Kermany) & Drusen, DME, CNV, Normal & OCT & 84,495 & N/A \\
Glaucoma Fundus & Glaucoma (Binary) & Fundus & 650 & 650 \\
Multimodal (OCT/A) & - & OCT & $>1,000$ & 2 \\
Yan et al. & - & Fundus & 18,394 & 5,825 \\
Cataract Dataset & Glaucoma, Cataract, Retinal Dis. & Fundus & 601 & 601 \\
OCT Challenge & Diabetic Retinopathy & OCT & 650 & 165 \\
\bottomrule
\end{tabular}
}
\end{table}

\subsection{Preprocessing Pipeline}
\begin{figure}[htbp]
    \centering
    \begin{minipage}{0.45\textwidth}
        \centering
        \includegraphics[width=0.7\linewidth]{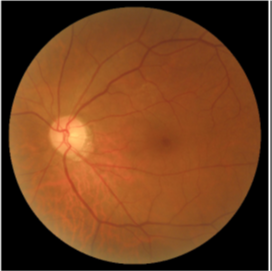} % The Original Image
        \caption*{A. Original Fundus Image}
    \end{minipage}\hfill
    \begin{minipage}{0.45\textwidth}
        \centering
        \includegraphics[width=0.7\linewidth]{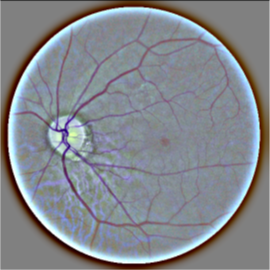} % The Graham's Method Result
        \caption*{B. Graham's Method Enhancement}
    \end{minipage}
    \caption{Visual comparison of the preprocessing pipeline. (A) The raw input image showing variable lighting conditions. (B) The output after applying Graham's method, revealing enhanced vascular contrast and normalized luminosity.}
    \label{fig:preprocessing}
\end{figure}

To reduce lighting variance and emphasize pathological features, we applied a rigorous preprocessing chain prior to network ingestion:

\begin{itemize}
    \item \textbf{ROI Cropping \& Margin Removal:} Raw fundus images often contain uninformative black margins. We implemented an automated contour detection algorithm to crop the image to the retinal Region of Interest (ROI).
    \item \textbf{Resolution Standardization:} All distinct inputs were resized to a uniform dimension (e.g., $224 \times 224$ or $299 \times 299$) to match the input requirements of the backbone architectures.
    \item \textbf{Graham’s Method (Luminosity Normalization):} We employed the enhancement technique proposed by Graham, which subtracts a Gaussian-blurred version of the image from the original. This operation normalizes luminosity across the retina, effectively highlighting vascular structures and lesions. The impact of this transformation is visualized in Figure \ref{fig:preprocessing}.
    \item \textbf{Data Augmentation:} To prevent overfitting, we applied random horizontal flips (simulating left/right eye mirroring) and random spatial shifts during the training phase.
\end{itemize}

\subsection{Deep Learning Architectures}
We evaluated five state-of-the-art Convolutional Neural Networks (CNNs) initialized with ImageNet weights: VGG16, VGG19, InceptionV3, ResNet50V2, and Xception.

\begin{figure}[htbp]
	\centering
	\includegraphics[width=0.95\linewidth]{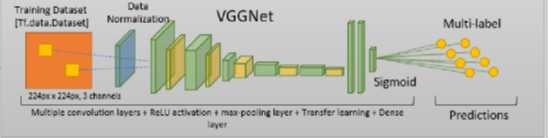}
	\caption{Schematic overview of the proposed VGGNet-based architecture. The pipeline integrates data normalization, deep feature extraction layers, and a multi-label classification head.}
	\label{fig:architecture}
\end{figure}

\paragraph{1. Feature Extraction}
We adopted a transfer learning approach, freezing the initial convolutional blocks to act as generic feature extractors. The fully connected classification heads were replaced with a custom top consisting of Global Average Pooling followed by a dense layer with 8 outputs (corresponding to the disease classes).

\paragraph{2. Hybrid SVM Classification}
To further refine classification boundaries, we explored a hybrid approach where the deep CNN serves solely as a feature embedding generator. The high-dimensional feature vectors extracted from the penultimate layer were fed into a Support Vector Machine (SVM) classifier. This combination leverages the feature learning power of Deep Learning with the margin-maximization properties of SVMs.

\subsection{Implementation Framework}
To ensure reproducibility and modularity, we developed a unified model interface.

\begin{itemize}
    \item \textbf{Model Abstraction:} We implemented a "BaseModel" wrapper class that standardizes the input/output interfaces across different backbones (VGG, ResNet, etc.).
    \item \textbf{Experiment Tracking (MLflow):} We integrated MLflow to systematically log hyperparameters, loss curves, and validation metrics.
\end{itemize}

% Placeholder for Fig 2 referenced in text
% \begin{figure}[ht]
% 	\centering
% 	\includegraphics[width=0.6\linewidth]{path/to/graham_method.jpg} 
% 	\caption{Effect of Graham's Method on retinal luminosity.}
% 	\label{fig:preprocessing}
% \end{figure}

\subsection{Deep Learning Architectures}
We evaluated five state-of-the-art Convolutional Neural Networks (CNNs) initialized with ImageNet weights: VGG16, VGG19, InceptionV3, ResNet50V2, and Xception.

\paragraph{1. Feature Extraction}
We adopted a transfer learning approach, freezing the initial convolutional blocks to act as generic feature extractors. The fully connected classification heads were replaced with a custom top consisting of Global Average Pooling followed by a dense layer with 8 outputs (corresponding to the disease classes).

\paragraph{2. Hybrid SVM Classification}
To further refine classification boundaries, we explored a hybrid approach where the deep CNN serves solely as a feature embedding generator. The high-dimensional feature vectors extracted from the penultimate layer were fed into a Support Vector Machine (SVM) classifier. This combination leverages the feature learning power of Deep Learning with the margin-maximization properties of SVMs.

\subsection{Implementation Framework}
To ensure reproducibility and modularity, we developed a unified model interface.

\begin{itemize}
    \item \textbf{Model Abstraction:} We implemented a "BaseModel" wrapper class that standardizes the input/output interfaces across different backbones (VGG, ResNet, etc.). This design pattern decouples the architecture from the training loop, allowing for seamless switching between backend frameworks (e.g., Keras to PyTorch) with minimal code refactoring.
    \item \textbf{Experiment Tracking (MLflow):} We integrated MLflow to systematically log hyperparameters, loss curves, and validation metrics. This allowed for granular comparison of model convergence and facilitated the selection of optimal hyperparameters.
\end{itemize}

\section{Explainability and Clinical Decision Support}
To mitigate the "black-box" nature of deep neural networks and facilitate human-in-the-loop diagnosis, the second phase of our framework focuses on interpretability. We propose two mechanisms to augment the clinician's assessment:

\subsection{Content-Based Image Retrieval (CBIR) via KNN}
To assist in differential diagnosis, we implemented a retrieval system that provides clinicians with historical reference cases similar to the patient under review. We utilize the feature embeddings extracted from the penultimate layer of our classification network to map input images into a high-dimensional latent space. A K-Nearest Neighbors (KNN) algorithm then identifies and retrieves the $k$ most semantically similar samples from the training corpus. This allows the physician to validate the model's prediction against confirmed historical diagnoses.

\subsection{Morphological Enhancement and Segmentation}
\begin{figure}[htbp]
	\centering
	\includegraphics[width=0.7\linewidth]{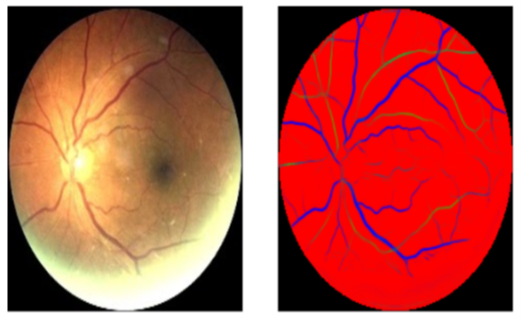} 
	\caption{Visualization of the Vessel Segmentation module. Left: The original fundus photograph. Right: The generated segmentation mask delineating the complex vascular tree, critical for analyzing hypertensive changes.}
	\label{fig:segmentation}
\end{figure}

Fundus photography projects 3D retinal structures onto a 2D plane, often obscuring subtle vascular details. To address this, we employ two enhancement strategies:

\begin{itemize}
    \item \textbf{Vessel Segmentation (W-Net):} Precise delineation of the retinal vasculature is critical for identifying hypertensive and diabetic anomalies. We utilize a W-Net architecture (two cascaded U-Nets) to generate high-fidelity binary vessel maps, filtering out background noise and highlighting vascular morphology. The output of this segmentation module is visualized in Figure \ref{fig:segmentation}.
    
    \item \textbf{Luminosity Normalization (Graham’s Method):} We provide the clinician with an enhanced view generated via Graham’s preprocessing technique. This method normalizes local luminosity variances, increasing the contrast of lesions and micro-aneurysms that may be missed in raw RGB images (refer to Figure \ref{fig:preprocessing} in Section VI).
\end{itemize}

\section{Experimental Results}
We evaluated the performance of the proposed architectures on the ODIR-5K and supplementary datasets. All models were trained using a batch size of 4 with a maximum of 20 epochs. To mitigate overfitting, we implemented an early stopping mechanism with a patience of 3 epochs, monitoring the validation loss. Empirical observation indicated that model convergence was typically achieved within 10 epochs.

\subsection{Baseline Model Comparison}
We benchmarked five architectures: VGG16, VGG19, ResNet50V2, InceptionV3, and Xception. The quantitative results for the training and validation sets are presented in Table \ref{tab:train_results} and Table \ref{tab:val_results}, respectively.

As evidenced by the metrics, the Xception architecture demonstrated the most robust generalization performance, achieving the highest validation accuracy ($86.60\%$) and AUC ($0.8435$). While ResNet50V2 achieved superior performance on the training set (Accuracy: $95.11\%$, Recall: $87.21\%$), it exhibited a significant generalization gap when applied to the validation set (Recall dropped to $35.59\%$), indicative of severe overfitting. Conversely, InceptionV3 failed to converge effectively on the validation set, yielding negligible precision and recall.

\begin{table}[htbp]
\centering
\caption{Performance Metrics on Training Set}
\label{tab:train_results}
\begin{tabular}{lccccc}
\toprule
\textbf{Model} & \textbf{Accuracy} & \textbf{AUC} & \textbf{Loss} & \textbf{Precision} & \textbf{Recall} \\ 
\midrule
VGG16 & 0.8939 & 0.9061 & 0.2444 & 0.6380 & 0.4716 \\
VGG19 & 0.8910 & 0.9046 & 0.2492 & 0.6201 & 0.4692 \\
ResNetV2 & \textbf{0.9511} & \textbf{0.9771} & \textbf{0.1249} & \textbf{0.7449} & \textbf{0.8721} \\
InceptionV3 & 0.8660 & 0.7993 & 0.3166 & 0.0337 & 0.4628 \\
Xception & 0.9320 & 0.9566 & 0.1691 & 0.6289 & 0.8190 \\ 
\bottomrule
\end{tabular}
\end{table}

\begin{table}[htbp]
\centering
\caption{Performance Metrics on Validation Set}
\label{tab:val_results}
\begin{tabular}{lccccc}
\toprule
\textbf{Model} & \textbf{Accuracy} & \textbf{AUC} & \textbf{Loss} & \textbf{Precision} & \textbf{Recall} \\ 
\midrule
VGG16 & 0.8326 & 0.7990 & 0.3813 & 0.3757 & 0.2374 \\
VGG19 & 0.8498 & 0.7942 & 0.4214 & 0.4527 & 0.1833 \\
ResNetV2 & 0.8403 & 0.7779 & 0.5965 & 0.3778 & 0.3559 \\
InceptionV3 & 0.8553 & 0.5603 & 16.1981 & 0.0000 & 0.0000 \\
Xception & \textbf{0.8660} & \textbf{0.8435} & \textbf{0.3380} & \textbf{0.5469} & \textbf{0.3535} \\ 
\bottomrule
\end{tabular}
\end{table}

\subsection{Ablation Studies: Preprocessing and Transfer Learning}
We conducted further experiments to evaluate the impact of transfer learning and the proposed preprocessing pipeline (Graham’s method).

\begin{itemize}
    \item \textbf{Impact of Transfer Learning:} The utilization of ImageNet-pretrained weights significantly accelerated convergence for VGG and Xception models. However, for ResNet50V2, transfer learning yielded marginal gains, suggesting that the domain shift between ImageNet and fundus imagery requires more extensive fine-tuning for residual architectures.
    
    \item \textbf{Efficacy of Preprocessing:} The application of Graham’s luminosity normalization had a marked impact on training dynamics. Specifically, it acted as an effective regularizer for the ResNet model. While the preprocessing step introduced a slight computational overhead, it successfully mitigated the overfitting ("overflowing") observed in the baseline experiments, resulting in improved validation accuracy. For the Xception model, the inclusion of geometric augmentations (random flips and shifts) further enhanced classification performance across both subsets.
\end{itemize}

Detailed graphical representations of these ablation studies, including convergence plots, are provided in the Supplementary Material.

\section{Conclusion}
In this work, we presented a comprehensive Deep Learning framework for the automated screening of ocular pathologies. By leveraging transfer learning with the Xception architecture, we demonstrated robust classification performance across six specific disease categories using widely accessible fundus imagery.

Beyond binary classification, our primary contribution lies in the integration of an interpretable Clinical Decision Support System (CDSS). We addressed the "black-box" limitations of standard CNNs by incorporating:

\begin{itemize}
    \item \textbf{Semantic Segmentation:} Utilizing a W-Net to delineate retinal vasculature, aiding in the assessment of microvascular health.
    \item \textbf{Content-Based Retrieval:} A KNN-based module that retrieves historically similar cases to valid diagnostic hypotheses.
\end{itemize}

Our results suggest that this human-in-the-loop approach not only minimizes diagnostic error but also holds significant promise for medical education. By visualizing disease probabilities alongside enhanced morphological features, the system serves as a valuable pedagogical tool for training novice ophthalmologists, bridging the gap between theoretical knowledge and visual diagnosis.

% \bibliographystyle{unsrtnat}
% \bibliography{references}  %%% Uncomment this line and comment out the ``thebibliography'' section below to use the external .bib file (using bibtex) .

% %%% Uncomment this section and comment out the \bibliography{references} line above to use inline references.
% % \begin{thebibliography}{1}

% % 	\bibitem{kour2014real}
% % 	George Kour and Raid Saabne.
% % 	\newblock Real-time segmentation of on-line handwritten arabic script.
% % 	\newblock In {\em Frontiers in Handwriting Recognition (ICFHR), 2014 14th
% % 			International Conference on}, pages 417--422. IEEE, 2014.

% % 	\bibitem{kour2014fast}
% % 	George Kour and Raid Saabne.
% % 	\newblock Fast classification of handwritten on-line arabic characters.
% % 	\newblock In {\em Soft Computing and Pattern Recognition (SoCPaR), 2014 6th
% % 			International Conference of}, pages 312--318. IEEE, 2014.

% % 	\bibitem{hadash2018estimate}
% % 	Guy Hadash, Einat Kermany, Boaz Carmeli, Ofer Lavi, George Kour, and Alon
% % 	Jacovi.
% % 	\newblock Estimate and replace: A novel approach to integrating deep neural
% % 	networks with existing applications.
% % 	\newblock {\em arXiv preprint arXiv:1804.09028}, 2018.

% % \end{thebibliography}

% --- REFERENCES ---
\setcitestyle{numbers,square}
\bibliographystyle{unsrtnat} % Use 'unsrtnat' for numbered citations [1], [2] sorted by appearance
\bibliography{references}    % The name of your .bib file (without extension)

\end{document}